# Toward Building Science Discovery Machines

**Abdullah Khalili[1*] and Abdelhamid Bouchachia[1]**

[1] Department of Computing and Informatics, Bournemouth University, United Kingdom.
[*] Correspondence: a.m.khalili@outlook.com

**Abstract** The dream of building machines that can do science has inspired scientists for decades. Remarkable advances have been made recently; however, we are still far from achieving this goal. In this paper, we focus on the scientific discovery process where a high level of reasoning and remarkable problem-solving ability are required. We review machine learning techniques used in scientific discovery with their limitations. We survey and discuss the main principles driving the scientific discovery process. These principles are used in different fields and by different scientists to solve problems and discover new knowledge. We provide many examples of the use of these principles in different fields such as physics, mathematics, and biology. We also review AI systems that attempt to implement some of these principles. We argue that building science discovery machines should be guided by these principles as an alternative to the dominant approach of current AI systems that focuses on narrow objectives. Building machines that fully incorporate these principles in an automated way might open the doors for many advancements.

## 1. Introduction

In [1-2] Penrose talked about the existence of three different worlds: the mental world, the physical world, and the mathematical world. The physical world is governed by laws that reside in the world of mathematics, our minds emerge from the physical world, and those minds are able to access the mathematical world by discovering mathematics, which is within the scope of reason.

In the mathematical world, Bourbaki [5] likened mathematics to a city, where the outlying districts and suburbs expand on the surrounding country. Plato believed that ideas or forms exist in some ideal world outside the physical world, which became later known as the 'Platonic world of forms' [3]. If Plato's realm exists, it is very unlikely that different parts of such realm are disconnected and do not have links with each other, they would be beautifully connected and one can navigate between different parts of that realm, and discover new hidden structures.

Although on rare occasions, the intellect might break through into those worlds and get a limited glimpse of those realms as described by Penrose [1], and illustrated through many examples by Hadamard [6], still in most times we follow certain procedures and principles to reconstruct those realms. Similar to the mathematical and physical worlds, a curtail aspect of the mental world and hence artificial intelligence is to build the maps that represent other realms by using a set of principles to reconstruct these original worlds and discover new knowledge. Today these principles represent the major driving force of the scientific discovery process. We argue that these principles should be used as guiding principles for building science discovery machines. The landscape of AI is extremely vast, in this paper we will focus on the scientific discovery process [6-8, 33-37, 41-42, 54-55, 77, 81, 85, 87-91, 116-123, 149-150].

The paper is organized as follows: Section 2 reviews machine learning techniques used in scientific discovery with their limitations. Section 3 discusses and reviews the main principles used by scientists to solve problems and discover new knowledge. We argue that in addition to analogy and concepts combination, other principles should be incorporated to automate the scientific discovery process. We provide many examples of the use of these principles in different fields such as physics, mathematics, and biology. We also list AI systems that attempt to implement some of these principles. Section 4 discusses and concludes the paper with future research directions.

## 2. Related Work

Bacon [91, 100] argued that scientific discovery is based on the application of logical rules to observations. Popper [91, 101] suggested that scientific discovery is based on a process of conjecture and falsification. Langley et al. [91, 102] argued for the importance of heuristic search to quickly find a good solution in comparison to random search. The performance of heuristic search based systems degrades when the search space is extremely large, which limits the applicability of these systems. Some remarkable systems under this category include AM [19] and Eurisko [20] which output helpful mathematical concepts. The success of many science discovery systems such as AM, Eurisko, and BACON depends largely on a specific and optimal encoding of knowledge [21, 94], which leaves the challenge of representation building open.

More recently, Sozou et al. [91] presented a survey on different scientific discovery systems, these systems can be classified into these main categories: rule-based reasoning systems, data mining based systems, evolutionary computation based systems, and automation of scientific experiments. They also surveyed the psychological and philosophical aspects of the scientific discovery process. Their conclusion was that the role of human scientists will remain essential for the foreseeable future.

Roscher et al. [87] provided a survey of recent applications of machine learning in scientific discovery and the way that explainable machine learning is used in combination with domain knowledge from the application areas. They reviewed recent uses of machine learning for molecular and materials science [103], physical sciences [104], biology and medicine [105, 106], and earth science [107]. Raghu et al. [99] presented a comprehensive guide to help scientists to use different deep learning methods in template ways for scientific discovery. They provided different science use cases for different deep learning methods such as Convolutional Neural Networks (CNNs), Graph Neural Networks (GNNs), Recurrent Neural Networks (RNNs), transformers, etc. They also reviewed new advances on using less labeled data such as self-supervision, semi-supervised learning, and data augmentation. Where in many scientific use cases only a small amount of data may be available to train machine learning models.

Deep learning models are very effective in pattern recognition, but they still have many limitations in high-level functions such as reasoning. Furthermore, they do not have the flexibility to generalize to new tasks, they are also vulnerable to adversarial samples [11-13, 80, 83, 86]. Yuan et al. [12] proposed an architecture that can produce images correctly classified by human subjects but misclassified by a deep network with a 97% adversarial success rate by only changing 4% of the image on average. Su et al. [13] showed that modifying one pixel only could lead up to 73% adversarial success rate depending on the used images. Recently, there is a growing interest in building neural networks that can learn to reason [76-79]. However, there is still a clear gap between neural networks and humans even when the machines have intensive training. The strongest objection to the deep learning approach is that it is unlikely to achieve general intelligence via self-organizing networks of neurons.

Other machine learning approaches such as logic-based approaches have remarkable representational power, logic is also crucial to achieve high-level functions such as reasoning. However, these approaches tend to be limited in learning and creativity, they are also limited in handling noise and uncertainty present in many applications. See [15, 92] for recent advances and different attempts to combine logical AI and neural networks. There is also a growing interest recently in studying artificial general intelligence [9, 10, 16-18, 85, 146-147]. Other important aspects of the scientific discovery process include explainable artificial intelligence to provide explanations of the discovery [128-129], and causal modeling [124] which is a key building block in the scientific discovery process.

## 3. The main principles driving the scientific discovery process

The dominant approach of current AI systems focus on solving specific narrow problems, this approach has key limitations in its generalization ability as discussed in earlier sections. The evolution-inspired path [83, 126-127] could provide an alternative way to build more general AI systems. However, the extremely large search space and the existence of many complex interacting parts still represent a major obstacle. In this study, we argue that the building of these systems should be guided by a set of principles as an alternative of narrow objectives or open-ended evolution. The use of these principles is backed by many historical examples of how different scientists made their discovery. Most scientific discoveries could be understood as instances of the use of one or more of these principles. These principles are the main approach used by scientists to solve problems and discover new knowledge. The use of these principles provides a way for machine learning systems to improve their generalization ability and to cut down the large search space of hypotheses by approaching the given problem using these principles in template ways as an alternative of the expensive random search.

In addition to logic, which plays an important role in the scientific discovery process, in reality, logic alone is not enough, we usually use more sophisticated principles and structures. In the literature, there is a focus on two main principles, concepts combination and analogies. However, other principles should be taken into account to build a comprehensive framework. Different problems in science can be solved using one or more of these principles by using these principles in template ways. For instance, some problems require finding the equation that fits the experimental data, some problems require finding the optimization criteria that give rise to the observed phenomenon, other problems require finding the rules or the program that gives rise to the observed phenomenon, many problems require combining different ideas, unifying ideas or finding analogy with other ideas, and so on. Each scientific problem comes with an objective to meet, the problem could be approached by these principles to find which principle best satisfy the objective. Proposing theoretical and computational frameworks that encapsulate these principles is beyond the scope of this paper. These principles can be summarized as follow

### 3.1 Mathematization

Mathematic is a very powerful tool in describing the natural world [30-32]. Mathematic today is very effective in studying fields as diverse as physics, computer science, finance, and biology. Mathematic is not only able to describe the natural world, but this description on many occasions led us to predict and discover new aspects of the studied phenomena. On many occasions, testing the mathematical description in new extreme conditions led to new insights and sometimes to new theories. In 1915 for instance, General Relativity (GR) was at the frontier of the map of physics, many physicists used the mathematization principle to derive new knowledge from the GR equation, they were able to predict gravitational waves, and black holes as solutions to the GR equation, both of these phenomena were confirmed experimentally in the few recent years.

In AI, there are many attempts to build symbolic regression algorithms, which are automated tools to find the mathematical equation that fits the experimental data [33]. Udrescu and Tegmark [34, 148] developed an algorithm that combines neural network fitting with a set of physics-inspired techniques. They applied it to 100 equations from the Feynman lectures on physics. It was able to discover all of them; the state of the art algorithm was only able to discover 71. For a more difficult test set, the state of the art success rate was improved from 15% to 90%. Many researchers recently [35-37] started to use recent advances in deep learning such as generative adversarial networks to discover physical concepts from experimental data without being provided with any additional prior knowledge and then use the discovered representation to answer questions about the physical system. The main limitation of these approaches is that it is difficult to transform the learned representations into interpretable properties unless prior knowledge is available. The main purpose of the algorithm that encapsulates the mathematization principle would be to find the equations that describe the experimental data.

## 3.2 Optimization

Optimization is one of the most powerful and most used principles from the least action principle in physics, survival of the fittest in biology, and utility maximization in economics, see [134-135] for a long list of examples from different disciplines. Optimization is one of the most used principles in everyday life, we constantly try to minimize energy, cost, distance, time, etc. Some other notable uses of this principle in science include minimizing the energy and time that are required to distribute fuels to the cells, gives rise to the circulatory system networks [27]. Optimizing the balance between the input and output energy gives rise to bird migration patterns [28]. Increasing entropy drives matter to acquire life-like physical properties [29]. The main purpose of the algorithm that encapsulates the optimization principle would be to find the optimization criteria and constraints that describe the studied problem.

## 3.3 Analogies

Many cognitive scientists [38] consider analogy to be one of the main building blocks of human cognition. There are many examples where analogy has played a crucial role in scientific discovery. Polya [39] observed that analogy has played a role in most mathematical discoveries. He provided many historical examples where analogy played the main role. See [40] for a long list of the use of analogy in scientific discovery. Nersessian [41-42] also gave a list of examples such as Newton's analogy between projectiles and the moon which gave rise to universal gravitation, Darwin's analogy between selective breeding and reproduction in nature which gave rise to natural selection, and the Rutherford-Bohr analogy between the structure of the solar system and the configuration of subatomic particles. Many algorithms in computer science have been inspired from biology to solve different problems such as the traveling salesman problem [43], [44]. They took inspirations from ants, which are capable of finding the shortest path from the nest to a food source [45], [46], by using a chemical substance called pheromone. Other notable examples include genetic algorithms, see [47-49] for a list of bio-inspired algorithms.

Two of the most remarkable approaches to this principle are the High-Level Perception (HLP) theory of analogy [94] and the Structure Mapping Theory (SMT) [22, 25], however;

building representations and models of the world still represents a major obstacle for these approaches. Hill et al. [93] investigated the use of neural networks to solve analogical problems, they also took inspiration from both SMT and HLP, where they encouraged the models to compare inputs at the more abstract level of relations rather than the less abstract level of attributes. Zhang et al. [26] also took inspiration from the field of psychology and education where teaching new concepts by comparing with noisy examples is shown to be effective. They build a model that sets the new state-of-the-art on two major Raven's Progressive Matrices datasets. One key limitation of the deep learning approach is the lack of transparency where the biases in many of the used datasets often lead to finding shortcuts instead of finding the real analogy [115, 132, 133]. See [22, 25, 93-96, 115] for different theoretical and computational frameworks for the analogy principle. The main purpose of the algorithm that encapsulates the analogy principle would be to find matching between the studied problem and similar problems.

### 3.4 Concepts Combination

Concepts Combination is a fundamental cognitive principle [50-52]. Many scientific discoveries are based on conceptual combination, where new concepts arise by combining old ones [53-55]. Concepts combination is also one of the main used themes in theoretical physics. In 1973 for instance, both general relativity and quantum mechanics were at the frontier of the map of physics, by combining ideas from these two fields, Hawking proposed that black holes emit thermal radiation. Moreover, by combining ideas from quantum mechanics and statistical mechanics, Bekenstein and Hawking proposed the formula that describes the black hole entropy, which later led to the holographic principle.

Some of the most notable approaches include conceptual blending [50], amalgamation [23], and compositional adaptation [24]. These techniques combine input concepts from a knowledge base and output novel concepts. See [23, 24, 50, 56-57, 113, 114] for different theoretical and computational frameworks. One major limitation of these techniques is that they require well-formed knowledge as input. Furthermore, without deeper representations and models of the world these approaches and other AI systems will keep operating at a very shallow level.

### 3.5 Emergence

Emergence is a powerful approach to explain complex behaviors by simple underlying rules. One notable example is birds flocking, some birds fly in coordinated flocks that show remarkable synchronization in movements. Heppner [60] showed that the coordinated movements could be the result of simple movement rules followed by each bird individually. Another example is the Game of Life [61], a two-dimensional cellular automaton with rules that avoid the formation of structures that grow freely or quickly disappear. Remarkable behaviors have been observed such as the glider, a small group of cells that moves like an independent emergent entity. Wolfram [62] used a cellular automaton with simple initial conditions and simple rules to produce highly complex behaviors. In [112] a new computational tool was introduced to model the emergence of more complex phenomena by allowing non-local and time-independent events to take place. More recently, Gregor et al. [125] proposed an artificial life framework to facilitate the emergence of intelligent organisms through evolutionary process. Graph neural networks could be more suitable than other techniques to model complex systems with multiple interacting parts. The main purpose of the algorithm that encapsulates the emergence principle would be to find the set of rules that gives rise to the emergent behavior.

## 3.6 Computability

Computation is a new paradigm that has revolutionized science and engineering [63, 82], it has derived many advancements in science and changed the way it is done. Many biologists would agree that biology is information science. One of the most notable examples is the DNA, which gives rise to the whole biological system. A growing number of physicists would also agree that the interactions between physical systems are information processing [64-65]. Zenil et al. [81] proposed a universal unsupervised and parameter-free model-oriented approach based on the concept of algorithmic probability to decompose an observation into its most likely algorithmic generative models. The approach uses a perturbation-based causal calculus and principles drawn from algorithmic complexity to infer model representations. They demonstrated the ability of the approach to deconvolve interacting mechanisms regardless of whether the resulted objects are bit strings, images, or networks. A related topic is using machine learning for code generation (see [98] for a recent survey). Finding the program, the rules, or the equation that underlies the studied phenomena would be an important step to improve the explainability of machine learning systems and to avoid shortcuts reported in many examples [132-133]. The main purpose of the algorithm that encapsulates the computability principle would be to find the program that gives rise to the observed phenomenon.

## 3.7 Beauty

Aesthetic judgments play a guiding role in scientific discovery [66-69]. Scientists often evaluate models and theories based on their aesthetic appeal.

The role of beauty in science has found some skepticism [71] because we still do not have a satisfactory theory that can exactly test the claims made by scientists about the beauty of a theory. Hume [108] argued that aesthetic appreciation is due to feelings. Baumgarten [109] took the view that aesthetic appreciation is objective. Kant argued that there is a universal aspect to aesthetic [110]. Dirac argued that beauty in mathematics is objective and universal [69, 70]. Recent works on empirical aesthetics [111] show that there is a general agreement on what is considered beautiful. A recent interesting study about the nature of aesthetic in science by Zeki et al. [72] demonstrated that the aesthetic appreciation of mathematical equations corresponds to the same brain activity that corresponds to the appreciation of music and art. Zee [73] and Thuan [74] also argued that beauty's attributes such as simplicity and symmetry have universal values and that they should not be subject to revision in science [69].

Recent approaches for beauty assessments of visual contents [130-131] could shed new lights on how to assess different scientific models. Deep learning could be particularly interesting where promising results were reported. The main purpose of the algorithm that encapsulates the beauty principle would be to find a metric that evaluates the scientific model describing the observed phenomenon.

## 3.8 Universality

Universality means that a similar mathematical formulation can describe different phenomena across multiple fields. The spectral measurements of composite materials, such as sea ice and human bones, the time between the buses' arrival in the city of Cuernavaca in Mexico, the zeros of the Riemann zeta function, and many other phenomena have shown to have the same statistical distribution [58]. Power laws are another example of universal laws that have been observed in a wide range of phenomena in fields as diverse as physics, biology, and computer science [59]. Recently, Mocanu et al. [144] were able to significantly reduce the number of parameters of deep learning models with no decrease in performance by enforcing a power law distribution.

### 3.9 Unification

Unification [151-152] has played a key role in physics since Newton who unified celestial and terrestrial mechanics, Maxwell who unified electricity and magnetism, then the unification of the weak and the electromagnetic forces, and most recently the attempts to unify all the four fundamental forces. Unification has also played an important role in biology [140-141]. In addition to several attempts to unify different machine learning approaches such as neuro-symbolic [15, 92], neuro-evolution [143], and many others [10, 14].

### 3.10 Symmetry

Symmetry has played an important role in science [75, 136-139] from Newton's laws to Maxwell's equations, and general relativity. Symmetry has also played a fundamental role in the development of quantum mechanics. Today, it is one of the most used principles in searching for the fundamental laws of physics and further unification. Convolutional neural networks represent an early use of the symmetry principle in deep learning. Recently, more advanced symmetry was used to significantly reduce the number of examples required to train deep learning models [142].

Many of these principles could operate at different levels, for instance, the circulatory system example in the optimization principle. By studying the literatures, one can find that the energy and time should be minimized; here the optimization principle is operating at the conceptual level. Then the principle could operate at the mathematical level by using a mathematical description of the optimization process. Similar reasoning could be applied for other principles such as concepts combination where the ideas are firstly combined at the conceptual level and then at the mathematical level.

The use of the above principles varies from field to field, some principles are still used in a limited fashion. For instant, using the computability principle is still very limited in physics, the use of mathematization principle is still very limited in social sciences, and the use of the beauty principle is more dominant in physics and mathematics than in biology.

There are many other domain-specific principles that are specific to certain fields. Finding new principles could be crucial to make new discoveries and revolutionize our understanding, for instance, the use of the symmetry and mathematization principles has revolutionized modern physics, and maybe we need new principles to see in new perspectives and solve current challenges.

### 4. Discussion and Conclusion

This paper has presented a review of different machine learning techniques used in scientific discovery with their limitations. It discussed and reviewed the main principles used by scientists to solve problems and discover new knowledge. We argue that a key step to improve the generalization ability of AI systems is to build systems guided by these principles rather than focusing on solving specific and narrow problems, or searching the extremely large space of the evolution-inspired approaches. The main challenge to build science discovery machines and automate the scientific discovery process is to build the theoretical and computational frameworks that encapsulate these principles. Although some principles are harder to automate where the challenge of building representation and models of the world is more dominant such as concepts combination and analogy. However, a lot of progress can be made in working on other principles such as mathematization, emergence, etc. Deep learning could be a very effective tool to implement some of these principles, it has shown promising results for the mathematization principle. However, it might be limited for other principles. In the literature, there is a focus on few principles, we believe that there are rooms for many interesting future contributions by working on the rest of the principles by building different theoretical and computational frameworks or by investigating the use of some existing AI techniques. Incorporating these principles fully in an automated scientific discovery framework might open the doors for many advancements. Pursuing this research direction holds a great promise to help scientist in their research and to speed up the scientific discovery process.